\documentclass{article}

\usepackage[eandd,final]{neurips_2026}

\usepackage[utf8]{inputenc}
\usepackage[T1]{fontenc}
\usepackage{hyperref}
\hypersetup{hidelinks}
\usepackage{url}
\usepackage{booktabs}
\usepackage{amsfonts}
\usepackage{amsmath}
\usepackage{amssymb}
\usepackage{nicefrac}
\usepackage{microtype}
\usepackage{xcolor}
\usepackage{graphicx}
\usepackage{multirow}
\usepackage{array}
\usepackage{enumitem}
\usepackage{placeins}

\newcommand{\sbs}{\textsc{SBS}}
\newcommand{\vbs}{\textsc{VBS}}
\newcommand{\headroom}{\Delta_{\text{route}}}
\newcommand{\tieset}{\mathcal{T}_i}
\newcommand{\sbsidx}{k_{\textsc{SBS}}}
\newcommand{\sbsscore}{S_{\textsc{SBS}}}
\newcommand{\vbsscore}{S_{\textsc{VBS}}}

\title{Routeability Before Routing:\\
Routeability Audit Protocol (RAP) and RouteabilityBench for Audited LLM Model Selection}

\author{%
  \begin{minipage}{0.96\textwidth}
  \centering\normalfont
  {\bfseries Yihang Lu}$^{1}$\qquad
  {\bfseries Denica Kjorvezir}$^{2}$\qquad
  {\bfseries Ana Gjorgjevikj}$^{2}$\\[0.65ex]
  {\bfseries Carola Doerr}$^{3}$\qquad
  {\bfseries Tome Eftimov}$^{2}$\\[1.2ex]
  {\small $^{1}$Sorbonne Universit\'e, CNRS, INRAE, IRD, iEES, Paris, France\\}
  {\small $^{2}$Computer Systems Department, Jo\v{z}ef Stefan Institute, Ljubljana, Slovenia\\}
  {\small $^{3}$LIP6, Sorbonne Universit\'e, CNRS, Paris, France\\}
  \vspace{0.35ex}
  {\small \texttt{yihang.lu@sorbonne-universite.fr}\\}
  \end{minipage}%
}

\begin{document}

\maketitle

\begin{abstract}
LLM routing aims to choose, for each prompt, which model in a portfolio is most likely to answer correctly. This only makes sense when the benchmark contains enough prompt-specific variation for different models to be best on different inputs. On coarse-scored LLM benchmarks, however, many models often tie for the best score, and the single best model may already leave little room for improvement. We therefore propose a routeability-first audit, instantiated as RAP and RouteabilityBench, a runnable claim-verification package. The audit is retrospective: it runs once the benchmark--portfolio score matrix exists and before router comparisons are interpreted as general evidence, rather than certifying a portfolio that has not yet been evaluated. Across four benchmarks and 66 LLMs, generic routing enhancements such as larger embedding sets, dimensionality reduction, selective abstention, clustering, and stronger regressors do not consistently outperform the single best model. The clearest positive evidence is concentrated in portfolio-dependent Omni-MATH settings, while GPQA, MMLU-Pro, and IFEval show different ways in which routing can remain hard or predictor-sensitive. Our practical recommendation is that routing gains should be reported together with routeability diagnostics, tied-best analysis, portfolio construction details, and family-corrected statistical tests.
\end{abstract}

\section{Introduction}
\label{sec:intro}

Per-prompt model routing asks a simple question: instead of using one large language model (LLM) for every prompt, can we choose a better model for each prompt from a portfolio? Recent work has explored this idea for both quality and cost, including routing benchmarks and learned routing systems~\citep{chen2024frugalgpt, ding2024hybridllm, ong2025routellm, shnitzer2024routing, hu2024routerbench}, and it now sits alongside broader LLM evaluation efforts such as HELM, MT-Bench, and Chatbot Arena~\citep{liang2023helm, zheng2023judging, chiang2024chatbotarena}. The appeal is obvious: if different models are strong on different prompts, routing could outperform using the same model everywhere. That promise depends on a hidden assumption. A useful router needs a benchmark where prompt-to-prompt differences actually matter for model choice. If the single best model is already close to an oracle upper bound, or if many models tie for the best score on the same prompt, then routing may be hard to learn no matter how sophisticated the router is. Our central question is therefore more basic than ``which router is best?'' Before comparing routers, we ask whether the benchmark and model portfolio define a meaningful routing problem at all. We call this property \emph{routeability}: whether the benchmark--portfolio pair contains enough learnable variation for routing to improve over always using the single best model.

To study that question, we package a routeability-first evaluation artifact: the \emph{Routeability Audit Protocol} (RAP) and its runnable evaluation package, \emph{RouteabilityBench}. RAP is a retrospective, pre-interpretation audit applied after the score matrix is available and before router comparisons are interpreted as general evidence; it is not a prospective certificate for an unevaluated portfolio. Portfolio-stratified auditing tests whether an apparent gain belongs to a particular benchmark--portfolio snapshot. The audit varies three axes: the dataset, the construction of the model portfolio, and the predictor family used to estimate prompt-level model performance. It also fixes the evaluation protocol through frozen 30-seed splits, train-only transformations, and family-corrected statistical tests~\citep{wilcoxon1945individual, holm1979simple}.

The main findings are simple to state. First, headroom alone is not enough: two benchmarks can offer similar oracle room for improvement while differing sharply in whether routing helps in practice. Second, common routing enhancements such as larger embedding sets, dimensionality reduction, abstention, clustering, or stronger regressors do not reliably outperform the single best model across all datasets. Third, the clearest positive evidence appears on Omni-MATH, while GPQA, MMLU-Pro, and IFEval each expose different kinds of boundary cases. The accompanying reviewer-runnable artifact packages frozen 30-seed split manifests, an authoritative filtered-row--prompt-ID registry, derived score/evidence tables, model and dataset metadata, statistical evidence tables, and one-command claim-verification scripts. It does not ship embeddings, prediction caches, or router-training code and therefore does not independently refit the routing models. We do not claim that LLM routing is impossible, nor do we propose a new universal router. Our goal is to audit when a benchmark and portfolio define a learnable routing problem. Positive results are interpreted narrowly: they identify settings where routing structure is detectable under controlled comparisons, not a general recipe for all LLM benchmarks.

\paragraph{Contributions.}
\begin{itemize}
\item[\textbf{C1}] \textbf{RAP: a routeability-first evaluation protocol.} We define and report SBS--VBS headroom, tied-best $|\tieset|$ distribution, regret, gap closure, selective-routing diagnostics, and named hypothesis families that should be audited \emph{before} citing routing gains (\S\ref{sec:diagnostics}).

\item[\textbf{C2}] \textbf{RouteabilityBench: a reviewer-runnable E\&D claim-verification artifact.} RouteabilityBench instantiates RAP with frozen 30-seed splits, a fail-loud prompt-ID alignment contract, derived score/evidence tables, model and dataset metadata, claim-level uncertainty summaries, and executable statistical checks (Appendices~A--J). The artifact verifies shipped evidence; full embedding pipelines and router refits are outside its release scope.

\item[\textbf{C3}] \textbf{Portfolio-stratified and predictor-stratified boundary findings.} We identify Omni-MATH as the clearest portfolio-sensitive positive setting, IFEval as a predictor-sensitive boundary case, GPQA as a high-headroom tied-best boundary, and MMLU-Pro as a tied-best / low-utility boundary. These are evaluation findings, not a new SOTA router (\S\ref{sec:full-portfolio}--\ref{sec:portfolio}).

\item[\textbf{C4}] \textbf{Reporting and release guidance for routing benchmarks.} We provide a routeability-centered reporting checklist, artifact documentation pattern, and leakage/contamination disclosure standard for future routing benchmarks (Appendix~\ref{app:reporting-checklist} and \S\ref{sec:limitations}).
\end{itemize}


\section{Background and Setup}
\label{sec:background}

\paragraph{Key definitions.}
We use a small set of terms throughout the paper. The \emph{single best solver} (\sbs) is the one model with the best average training performance in the current portfolio. The \emph{virtual best solver} (\vbs) is an oracle that picks the best-scoring model separately for each prompt after seeing the true score. \emph{Headroom} is the gap between \vbs{} and \sbs{}, so it measures the maximum possible average gain from perfect routing. The \emph{tied-best set} for a prompt is the set of all models that share the top score on that prompt. \emph{Forced routing} must choose a model for every prompt, while \emph{selective routing} may abstain or fall back to \sbs{} when uncertainty is high. Throughout the paper, an \emph{audit} means a controlled comparison designed to test whether an apparent routing gain is robust to a specific confound.

Let $Y \in \mathbb{R}^{N \times K}$ denote the performance matrix of $K=66$ LLMs on $N$ prompts. For each prompt $i$, let $X_i^v \in \mathbb{R}^{d_v}$ denote its embedding under view $v$. Let $\sbsidx = \arg\max_k |\mathrm{train}|^{-1}\sum_{i \in \mathrm{train}} Y_{i,k}$ be the single-best-solver model index estimated on the training split, and let $\sbsscore = N_{\mathrm{test}}^{-1}\sum_{i \in \mathrm{test}} Y_{i,\sbsidx}$ be its test-set score. We write $\vbsscore = N_{\mathrm{test}}^{-1}\sum_{i \in \mathrm{test}} \max_k Y_{i,k}$ for the virtual-best-solver upper bound and define routing headroom as $\headroom = \vbsscore - \sbsscore$. The tied-best set at prompt $i$ is $\tieset = \{k : Y_{i,k} = \max_{k'} Y_{i,k'}\}$. Forced routing regret is $r_i^{\text{forced}} = \max_k Y_{i,k} - Y_{i,\pi(i)}$ for routing policy $\pi$. Within a split, gap closure is $1-\bar r^{\text{forced}}/\headroom$ whenever $\headroom>0$; reported 30-split summaries average these splitwise ratios. Throughout the paper, \sbs{} names the baseline concept, $\sbsidx$ the model index, and $\sbsscore$ the scalar score.

The audit retains four HELM-style single-turn benchmarks~\citep{liang2023helm}: GPQA ($446$ prompts), IFEval ($541$), MMLU-Pro ($1000$), and Omni-MATH ($1000$), for $2987$ score-matrix rows. The compact three-view routing analyses use the exact finite prompt-ID intersection: $445$, $541$, $1000$, and $998$ rows, respectively. Frozen split integers index this filtered order, not the unfiltered score matrices. The artifact therefore includes a 2,984-row registry mapping every filtered index to its prompt ID and original score row; its verifier requires exact ID matching, in-range unique indices, and complete coverage for all 120 dataset--seed partitions. WildBench was left out because its multi-turn conversational structure changes the representation target from standalone prompts to dialogue trajectories, and its contexts are much longer. Over the retained corpus, prompt length has mean $161.0$, median $134.8$, 95th percentile $403.3$, and maximum $2278.5$ tokens.

We use a fixed 66-LLM portfolio. The source-backed weight-access taxonomy contains 22 open-weight, 8 gated-weight, and 36 closed-weight models; the binary sensitivity analysis compares 22 open-weight with 44 gated-or-closed-weight models. Weight access is recorded for all 66 models, service access for 38, and capability tier is not recorded. All routeability quantities, \sbs/\vbs{} baselines, and portfolio-construction tests are conditional on this versioned portfolio snapshot unless a section explicitly restricts the model set. Prompt features come from several embedding views. We use three compact encoder-style views (BGE, E5, GIST), three larger decoder-based views (Qwen3-8B, Qwen3-4B, Nemotron-8B) with and without chunking, their six-view combinations, and a full nine-view combination. All views are converted into fixed-length, L2-normalized prompt vectors using attention-mask-aware mean pooling. For chunked inputs, overlapping windows are aggregated by token-count-weighted averaging; tokens in overlaps can therefore receive repeated weight, so this is a documented approximation rather than exact equal weighting of original tokens. When PCA is used, it is always fit on training rows only. Appendix~\ref{app:dataset-embedding} gives the full view table, dimensions, pooling, chunking, and quality checks.

All experiments use the frozen 30-seed train/validation/calibration/test split manifest, with approximately 50/15/15/20 proportions subject to integer rounding. Original seed-level screens used paired Wilcoxon signed-rank tests with within-family Holm correction; corrected primary claims use prompt-cluster bootstrap intervals and prompt-level sign-flip tests with separately registered Holm families, while overlapping-split Nadeau--Bengio analysis is a sensitivity check~\citep{nadeau2003inference, demsar2006statistical, dietterich1998approximate, benavoli2017time}. These repeated splits quantify split-induced variability; they are not 30 independent benchmark datasets. Routed actions use \texttt{argmax\_random\_tiebreak} to avoid alphabetic bias under tied score vectors. Ridge regularization $\lambda$ is selected by 5-fold cross-validation strictly on the training split. The accompanying artifact packages the split and prompt-ID manifests, derived score/evidence tables, metadata, and executable claim checks. It does not distribute prompt features, frozen predictions, or training code; refitting routers is outside the artifact's reproduction scope.

\paragraph{Connection to Algorithm Selection.}
LLM routing is a form of per-instance algorithm or model selection in the sense of Rice's classical formulation: we observe instance features, choose among a portfolio of candidate solvers, and care about the resulting performance matrix rather than the average performance of any single solver~\citep{rice1976algorithm}. This framing connects the paper to the automated algorithm selection literature, including SATzilla, ASlib, and AutoFolio~\citep{xu2008satzilla, bischl2016aslib, lindauer2015autofolio}, as well as broader surveys of per-instance selection~\citep{kerschke2019automated}. The analogy is useful but incomplete. In classical AAS settings, the instance-feature representation and algorithm-performance matrix are often comparatively stable, whereas LLM routing must work with semantic prompt features, coarse or tied score matrices, and portfolios whose composition can dominate downstream utility. Our routeability diagnostics are therefore designed as a front-end audit: before importing AAS-style routing claims, we first ask whether the benchmark, score granularity, and portfolio induce a learnable decision problem at all.

\paragraph{Reader map.}
The rest of the paper follows the audit in the order a reviewer would naturally ask questions. Section~\ref{sec:diagnostics} asks whether routing is plausible in principle by reporting headroom and tied-best diagnostics. Section~\ref{sec:full-portfolio} asks whether simple full-portfolio routers outperform the single best model. Section~\ref{sec:dim} asks whether the answer changes under different embedding views, dimensionality controls, or predictor families. Section~\ref{sec:portfolio} asks whether changing the portfolio itself reveals routing structure that the full 66-model setting hides. The final sections summarize the cost caveat, reporting recommendations, and bounded conclusions.

\section{Routeability Diagnostic Suite}
\label{sec:diagnostics}

Before interpreting router comparisons, we ask whether the benchmark--portfolio snapshot supplies a meaningful routing problem. A structural diagnostic can reveal headroom, tied-best saturation, or calibration stress, but it cannot guarantee that a learned router will beat the SBS. This retrospective screen therefore separates structural opportunity from realized routing utility.

We use five diagnostics. D1 is SBS/VBS headroom, $\headroom=\vbs-\sbs$, the maximum average improvement available to a perfect selector. D2 is tied-best saturation, summarized by the distribution of $|\tieset|$, which measures how often many LLMs share the top score. D3 measures embedding-to-performance predictability using row-wise ranking or correlation diagnostics between prompt embeddings and LLM performance vectors. D4 is realized decision utility, reported through regret and gap closure for forced routing policies. D5 covers calibration-sensitive selective routing: coverage, threshold transfer, and abstention behavior. Together these diagnostics define the regime in which routing claims should be interpreted.

Table~\ref{tab:diagnostics} separates four routeability patterns. GPQA and Omni-MATH have similar headroom, but only Omni-MATH later yields retained, protocol-specific portfolio cells; the difference is not the oracle ceiling alone, but also score granularity and tied-best mass. These cells share a score matrix and are not independent replications. MMLU-Pro combines lower headroom with binary scoring and heavy tied-best saturation. IFEval combines low headroom with the heaviest tied-best saturation, but later predictor audits show a boundary effect. Headroom is a ceiling, not a promise of learnable routing.

\begin{table}[htbp]
  \caption{Routeability diagnostics across four benchmarks ($K{=}66$). These SBS/VBS values are full-matrix descriptive diagnostics; train/test comparisons select the SBS on training rows and evaluate it on held-out rows. Tied-best fraction is $|\tieset|/K$.}
  \label{tab:diagnostics}
  \centering
  \small
  \begin{tabular}{lccccccc}
    \toprule
    Dataset & $N$ & $\sbs$ & $\vbs$ & $\headroom$ & mean $|\tieset|$ & median $|\tieset|/K$ & score levels \\
    \midrule
    Omni-MATH & 1000 & 0.722 & 0.922 & \textbf{0.200} & 27.3 & 36\% & 5 \\
    GPQA      & 446  & 0.803 & 0.996 & \textbf{0.193} & 34.6 & 55\% & 2 \\
    MMLU-Pro  & 1000 & 0.903 & 0.994 & 0.091 & 44.4 & 76\% & 2 \\
    IFEval    & 541  & 0.951 & 0.996 & 0.045 & 49.3 & 88\% & 5 \\
    \bottomrule
  \end{tabular}
\end{table}

Coarse scoring creates large tied-best sets, so strict winner-label classification can become unstable even when regret-based routing remains meaningful. In three of the four benchmarks, the median tied-best fraction exceeds half the portfolio. Under such saturation, a top-1 selector must choose one arbitrary representative from many score-equivalent LLMs; small perturbations in tie-breaking or labels can change strict accuracy without changing utility. We therefore use full performance-vector prediction and tie-aware regret/gap-closure metrics rather than treating strict single-winner accuracy as the primary evidence of routeability.

\paragraph{Proposition 1 (Strict-winner labels are unstable under tied-best saturation).}
For prompt $i$, let $\tieset=\{k:Y_{i,k}=\max_{k'} Y_{i,k'}\}$ be the tied-best set. Suppose an evaluation converts $\tieset$ into a single strict winner label $Z_i$ by choosing uniformly from $\tieset$. Any policy $\pi$ that selects a top-scoring model, $\pi(i)\in \tieset$, has zero tie-aware regret on prompt $i$, but expected strict-label accuracy $1/|\tieset|$. Appendix~\ref{app:strict-winner-proof} gives the short proof. This proposition explains why RAP treats strict winner accuracy as a side diagnostic and centers tie-aware regret and gap closure.

\paragraph{Mechanism diagnostic: headroom, score granularity, and tied-best mass.}
A lightweight mechanism diagnostic supports this interpretation. GPQA and Omni-MATH have similar SBS--VBS headroom ($0.193$ vs.\ $0.200$), but GPQA has binary scoring, heavier tied-best saturation (median $|\mathcal{T}_i|/K=54.5\%$), and only $1.1\%$ unique-best prompts. Omni-MATH combines high headroom with five-level scoring, lower tied-best saturation (median $36.4\%$), and the largest unique-best mass among the retained benchmarks ($8.4\%$). Thus, headroom alone does not characterize routeability; score granularity and tied-best mass also affect whether prompt-level model choice can be learned.

\paragraph{Hypothesis families and multiplicity.}
Table~\ref{tab:stat-families} registers the decision-relevant families. Families are corrected separately rather than pooled globally, so the family and inference unit are part of each claim.

\begin{table}[htbp]
  \caption{Hypothesis-family registry. Holm correction is applied within each stated family, never pooled. \emph{Tests} = seed-level screen of the original submission / family of the quoted prompt-level Holm value (H1: four-test prompt-level family; H2, H5, H6, H7: nine-test prompt-level headline family frozen on 2026-07-12; H4: ten-test original-score family; ``--'': none). Appendix~\ref{app:prereg-sensitivity} gives family-membership sensitivity.}
  \label{tab:stat-families}
  \centering
  \footnotesize
  \setlength{\tabcolsep}{2pt}
  \begin{tabular}{@{}>{\raggedright\arraybackslash}p{0.055\linewidth}>{\raggedright\arraybackslash}p{0.225\linewidth}>{\raggedright\arraybackslash}p{0.15\linewidth}>{\raggedright\arraybackslash}p{0.51\linewidth}@{}}
    \toprule
    ID & Family & Tests & Camera-ready decision \\
    \midrule
    H1 & Selective vs forced & 40 / 4 & four directional prompt-level comparisons survive Holm \\
    H2 & SBS dominance & 12 / 9 & Omni-MATH forced downgraded; prompt-level Holm $p=0.0912$ \\
    H3 & Weight-access sensitivity & 40 / -- & split-level Wilcoxon--Holm over 30 overlapping splits, $p_{\rm Holm}=0.004103$ \\
    H4 & Portfolio estimands & 56, 48 / 10 & prompt-space full-test survives in the original-score family only; LLM-side result reverses under common oracle \\
    H5 & PCA & 24 / 9 & PCA-64 withdrawn; 95\% interval includes zero \\
    H6 & Regressor family & 12 / 9 & IFEval ExtraTrees survives, sign-flip Holm $p=0.0126$ \\
    H7 & RF sensitivity & 4 / 9 & computationally unstable; no retained positive claim \\
    \bottomrule
  \end{tabular}
\end{table}

\section{Full-Portfolio Routing Baselines}
\label{sec:full-portfolio}

The baseline router predicts the full 66-dimensional performance vector from prompt embeddings using Ridge regression. For each view, $\lambda$ is selected by train-only 5-fold cross-validation; view predictions are combined using non-negative validation-learned fusion weights. Forced routing then selects the predicted best model with random tie-breaking. This baseline is intentionally simple: it gives a leakage-controlled, tie-aware reference point before the modeling-choice audits in \S\ref{sec:dim}.

Full-portfolio forced routing does not uniformly close the SBS--VBS gap. Across 30 seeds, forced gap closure (mean $\pm$ sample SD of splitwise ratios) is negative on GPQA ($-0.135\pm0.126$), MMLU-Pro ($-0.147\pm0.191$), and IFEval ($-0.324\pm0.477$), but positive on Omni-MATH ($+0.081\pm0.089$). This divergence is the first decision-level boundary: embedding/performance predictability can help estimate performance vectors, but it does not automatically translate into routing that outperforms the SBS.

The $K=66$ row of Table~\ref{tab:portfolio-size} reports routed-regret-minus-SBS-regret as mean $\pm$ sample SD across the 30 frozen repeated splits. These SDs and the gap-closure SDs above describe split-induced variability, not uncertainty across independent datasets; this frozen-prediction masking analysis is distinct from the operating-point family and does not refit a router per sub-portfolio.

Selective routing exposes calibration regimes rather than providing a generic fix. In prompt-level re-analysis, risk control lowers regret relative to forced routing on GPQA, IFEval, and MMLU-Pro, while Omni-MATH reverses direction. All four ordinary 95\% prompt-cluster bootstrap intervals exclude zero in the prespecified direction; separately, the four sign-flip tests survive Holm correction (adjusted $p=0.0436,0.0222,0.0222,0.0040$). The intervals are not multiplicity-adjusted. SBS dominance is stricter. After aligning the SBS comparison to the router's prompt domain, the seed-level screen is favorable for two Omni-MATH operating points, but the 30 splits reuse prompts. For forced routing, prompt-level re-analysis gives $\Delta=-0.0136$ with 95\% interval $[-0.0251,-0.0020]$, while the family-level sign-flip Holm value is $0.0912$. We therefore downgrade the SBS-dominance claim and make no cross-dataset dominance claim.

Appendix Table~\ref{tab:app-claim-uncertainty} defines the uncertainty-reporting fields used for these headline claims, including effect sizes, seed counts, and confidence-interval availability.

We also treat cross-dataset calibration transfer cautiously. In the corrected comparisons, the six displayed calibration-transfer tests are not Holm-significant; the smallest corrected value is GPQA$\rightarrow$Omni-MATH with $p_{\text{Holm}}=0.0624$. The safe interpretation is that selective thresholds appear dataset-specific under the corrected protocol, not that a dataset-agnostic threshold failure has been established.

RAP also adds a baseline robustness layer around this reference router. The internal evaluation suite includes four baseline classes, defined in Appendix~\ref{app:regressor-family} and Appendix~\ref{app:routellm}: trivial and near-trivial selectors (uniform random, frequency-matched random, train-mean selection), compact single-view and multiview baselines, kNN retrieval routing, and the audited full-vector regression family. The minimal candidate ships their frozen evidence but not the code and features needed to refit every baseline. These checks matter for an E\&D artifact because a benchmark can appear routeable relative to a weak router while still failing to improve over the simplest stable baselines.


\section{Modeling Choices Modulate but Do Not Universalize Routing}
\label{sec:dim}

The preceding results leave an important reviewer concern: perhaps the negative and boundary cases are artifacts of an insufficient representation, an overparameterized feature space, or a weak regressor. We therefore audit three modeling choices that are often treated as routine implementation details: embedding views, train-only dimensionality reduction, and regressor family. The audit changes the story in useful ways, but not into a broad routing win: modeling choices modulate regret in specific regimes, yet none yields routing that broadly outperforms the SBS across datasets.

\subsection{Representation and dimensionality controls}
The first audit asks whether richer prompt representations---that is, adding more embedding views beyond the compact encoder views---or train-only dimensionality control can rescue full-portfolio routing. Across the view-configuration family (compact encoder views, decoder-based views, and their concatenations), no representation setting provides a consistent routing gain after Holm correction. This is a boundary result rather than an indictment of representations: richer views may shift individual operating points, but adding larger or more views is not sufficient to create routeability where the benchmark--portfolio pair does not support it.

The train-only PCA audit addresses a different concern: high-dimensional embeddings might obscure routeability under Ridge. PCA is fit only on $X_{\text{train}}$ and then applied to validation, calibration, and test rows, preventing test-set leakage. The original seed-level screen produced one MMLU-Pro / compact-three-view / PCA-64 cell in a 24-test family. Prompt-level re-analysis reduces the effect to $\Delta=-0.0017$ with 95\% bootstrap interval $[-0.0051,0.0018]$. We therefore withdraw the PCA-64 finding; the dimensionality audit provides no confirmed routing improvement.

\subsection{Regressor-family audit identifies IFEval as predictor-sensitive}
Regressor family is the clearest case where the audit revises the earlier Ridge-only interpretation. ExtraTrees reduces IFEval forced-routing regret relative to Ridge under prompt-level inference ($\Delta=-0.0164$, 95\% bootstrap interval $[-0.0266,-0.0063]$, family-level sign-flip Holm $p=0.0126$), and the direction survives the audited software environments and 50 tie-break realizations. RandomForest points in the same direction in some realizations, but its original prompt-level interval reaches zero ($[-0.0233,0.0002]$); we therefore classify it as computationally unstable rather than as a positive finding. IFEval is a predictor-sensitive boundary regime, but this predictor comparison does not by itself show that either predictor beats the SBS.

Appendix Table~\ref{tab:app-claim-uncertainty} defines the uncertainty-reporting fields used for the PCA, ExtraTrees, and RandomForest evidence, including raw $p$-values, Holm-adjusted $p$-values, effect sizes, and confidence-interval availability.

\subsection{Implication}
These modeling-choice audits explain why a single router, embedding package, dimensionality choice, or predictor family is not an adequate unit of evidence for LLM routing. Section~\ref{sec:portfolio} shifts to portfolio construction, where the interpretation depends on the comparator and regret oracle. Positive cells are reported separately rather than treated as independent converging evidence.

\begin{figure}[t]
  \centering
  \includegraphics[width=\linewidth]{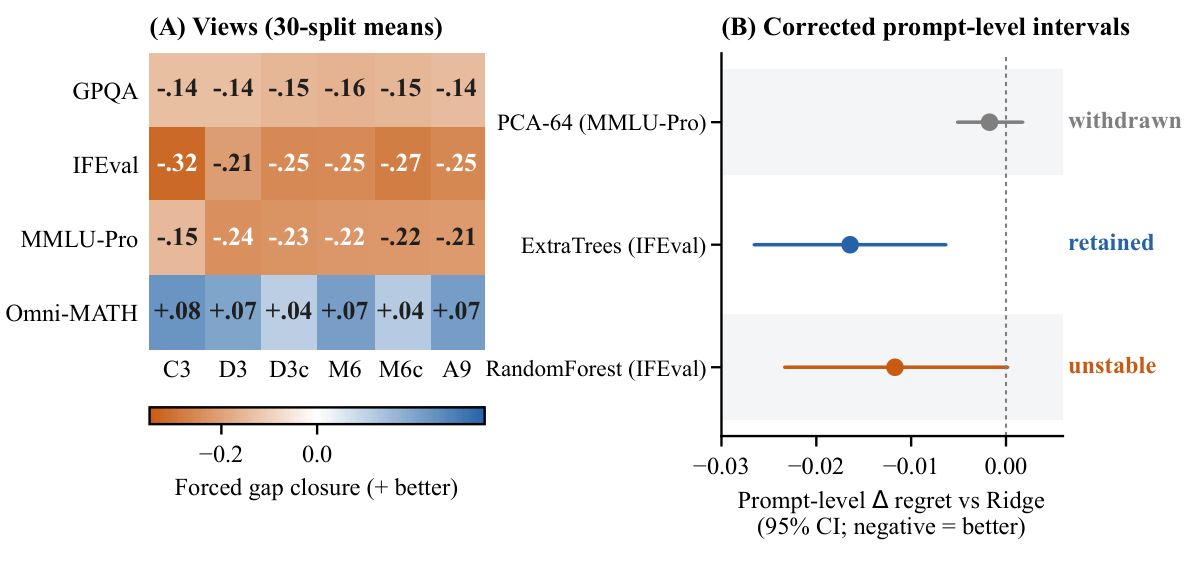}
  \caption{Modeling-choice audit. \textbf{(A)} Forced gap closure across embedding views remains dataset-dependent. C3 denotes the compact three views; D3/D3c the decoder three views without/with chunking; M6/M6c the mixed six views without/with chunking; and A9 all nine views. \textbf{(B)} Prompt-level 95\% intervals withdraw PCA-64, retain ExtraTrees on IFEval, and classify RandomForest as unstable because its interval reaches zero.}
  \label{fig:modeling-choices}
\end{figure}

\section{Portfolio Construction Is Estimand-Dependent}
\label{sec:portfolio}

We now move from modeling choices to portfolio construction. The access split, prompt-space conditioning, and LLM-side representative selection all produce informative Omni-MATH cells, but they share the same score matrix and estimate different quantities. We therefore report them as separate, protocol-specific analyses. Regret against each sub-portfolio's own oracle is not directly comparable to regret against a common full-portfolio oracle.

\subsection{The original full-portfolio clustering protocol is ill-posed}
\label{sec:original-cluster}
The first clustering audit is a useful negative control. A naive full-portfolio prompt-clustering protocol across the four datasets yields strict selector accuracy at most $0.013$, maximum configuration-level mean embedding-to-oracle ARI of $0.058$, and routable-cluster fraction approximately zero. We interpret this as a boundary-discovery result rather than a verdict against clustering in general. In the full 66-model setting, tied-best saturation makes strict winner labels structurally brittle, the per-cluster sample size is too small for stable cluster-level routing decisions, and semantic prompt geometry is only weakly aligned with the LLM performance space that determines regret.

\subsection{Weight-access sensitivity: taxonomy-dependent evidence on Omni-MATH}
\label{sec:access-group}
We reviewed all 66 models using source-backed metadata and recorded weight access separately from service access (known for 38 models) and capability tier (not recorded). The binary audit compares open-weight ($22$) with gated-or-closed-weight ($44$) portfolios and evaluates five prespecified routing methods per dataset and group, so Holm correction covers the complete $4\times2\times5=40$ family. One cell has a split-level Wilcoxon--Holm value below $0.05$: Omni-MATH margin abstention in the gated-or-closed portfolio ($p_{\text{Holm}}=0.004103$; mean regret difference relative to its group SBS $=-0.008458$). This result is calculated over 30 overlapping splits; the released evidence does not provide an independent prompt-level or Nadeau--Bengio confirmation for the revised 22/44 grouping. Because attribution changes under plausible taxonomies, it supports portfolio composition as a versioned sensitivity axis, not an intrinsic ordering of access tiers.

\paragraph{Portfolio-size and deletion sensitivity.}
We mask the frozen prediction matrix to uniform sub-portfolios at five completed sizes ($K=7,13,33,50,66$), use provider-family-stratified draws at $K=13,33,50$, and perform leave-one-out deletions. Structural diagnostics were additionally swept at $K=20,59$; routed-utility masking was not, so those sizes are not in Table~\ref{tab:portfolio-size}. Omni-MATH becomes favorable relative to the fixed full-66 SBS reference in the median at $K=33$ and strengthens with size; the other datasets remain unfavorable at the five completed uniform sizes. At small $K$ the sub-portfolio's own SBS is weaker and easier to beat (IFEval beats its own SBS in 56\% of $K=7$ draws), so Table~\ref{tab:portfolio-size} anchors on the fixed full-66 reference. No single-model deletion changes a regime label, while whole-provider-family removal changes IFEval when xAI is removed and Omni-MATH when OpenAI is removed. This supports local robustness, not invariance to arbitrary future portfolios.

\begin{table}[t]
  \caption{Routed-regret difference relative to the fixed full-66 SBS reference under uniform sub-portfolio masking. Negative favors routing. Rows $K<66$ are medians over 50 draws; $K=66$ is mean $\pm$ sample SD over the 30 frozen splits. Frozen predictions are masked without refitting.}
  \label{tab:portfolio-size}
  \centering
  \small
  \begin{tabular}{rcccc}
    \toprule
    $K$ & GPQA & IFEval & MMLU-Pro & Omni-MATH \\
    \midrule
    7 & $+.065$ & $+.019$ & $+.045$ & $+.067$ \\
    13 & $+.052$ & $+.015$ & $+.040$ & $+.002$ \\
    33 & $+.030$ & $+.011$ & $+.033$ & $-.006$ \\
    50 & $+.022$ & $+.011$ & $+.012$ & $-.014$ \\
    66 & $+.0242\pm.0228$ & $+.0137\pm.0189$ & $+.0127\pm.0172$ & $-.0164\pm.0179$ \\
    \bottomrule
  \end{tabular}
\end{table}

\subsection{LLM-side performance-profile clustering: internal gap closure versus deployment score}
\label{sec:llm-cluster}

We next ask whether we can build a smaller, more routeable portfolio by clustering models according to how they perform across prompts, rather than simply choosing the strongest models by average score. This matters because a portfolio made only of the strongest models may be too homogeneous: if those models tend to succeed on the same prompts, the router sees little useful diversity even though the models are individually strong. We compare centroid-medoid representatives from train-only performance-profile clustering against two controls: strength-only top-$R$ selection and random representative selection. The full audit spans four datasets, 30 seeds, and $K\in\{8,16,32\}$ representatives.

At Omni-MATH $K=8$, centroid medoids close more of their own subset-specific oracle gap than top-$R$: prompt-level re-analysis gives $\Delta=-0.0224$ with 95\% interval $[-0.0397,-0.0051]$ ($p_{\rm Holm}=0.0256$ in the ten-test original-score family, $0.115$ in a 30-test family that adds two binarized score versions; Appendix~\ref{app:prereg-sensitivity}); comparison with random representatives is null under that estimand. Against a common full-portfolio oracle, however, centroid medoids have \emph{higher} regret than top-$R$ ($\Delta=+0.0800$, $[0.0669,0.0937]$). Portfolio construction changes both the restricted problem's learnability and its attainable ceiling.

\subsection{Prompt-space cluster-conditioned routing survives a portfolio and label control}
\label{sec:prompt-cluster}
We next test whether routing becomes more useful after conditioning decisions on prompt clusters rather than treating the whole benchmark as one undifferentiated pool. If the best model depends on prompt type, then routeability may appear weak at the full-benchmark level while becoming clearer inside meaningful prompt groups. We use the compact three-view encoder representation to build train-only prompt clusters, then apply a per-cluster top-$M$ routing rule. To check that any gain is not an artifact of portfolio restriction or arbitrary labels, we compare this rule against two controls: a global top-$M$ baseline and a permuted-cluster-label baseline.

On Omni-MATH at $K=16,M=8$, prompt-level full-test re-analysis gives $\Delta=-0.0160$ versus global top-$M$ (95\% interval $[-0.0269,-0.0051]$) and $-0.0176$ versus permuted labels ($[-0.0265,-0.0088]$). Both survive correction within the 10-test original-score family. The global-top-$M$ contrast depends on score granularity: it is null when scores are binarized at $1.0$ ($\Delta=+0.0010$, $[-0.0127,+0.0150]$) and not Holm-significant at threshold $0.5$ ($p_{\rm Holm}=0.135$), and in a 30-test family that pools the three score versions its full-test $p_{\rm Holm}$ is $0.042$ (permuted labels: $0.005$; Appendix~\ref{app:prereg-sensitivity}). The operating point and routing definitions were frozen before the reconciliation outputs, but were inherited from the submitted analysis rather than selected on an independent held-out study; we therefore treat the result as protocol-specific, not as prospective confirmation or a universal clustering claim.

\begin{table}[t]
  \caption{Protocol-specific Omni-MATH portfolio results. Signs depend on the declared estimand; rows are not independent replications.}
  \label{tab:portfolio-summary}
  \centering
  \scriptsize
  \resizebox{\linewidth}{!}{%
  \begin{tabular}{p{0.22\linewidth}p{0.31\linewidth}p{0.24\linewidth}p{0.17\linewidth}}
    \toprule
    Protocol & Estimand / comparator & Result & Reading \\
    \midrule
    Weight access & group regret vs group SBS & one corrected 40-test cell & taxonomy-sensitive \\
    Prompt-space & full-test vs global top-$M$ & $-0.0160$ $[-.0269,-.0051]$ & retained \\
    LLM-side & own-oracle vs top-$R$ & $-0.0224$ $[-.0397,-.0051]$ & internal gap \\
    LLM-side & full-oracle vs top-$R$ & $+0.0800$ $[.0669,.0937]$ & reverses \\
    \bottomrule
  \end{tabular}
  }
\end{table}

The prompt-space comparison survives on the full test set, whereas the favorable LLM-side comparison uses a portfolio-specific oracle and reverses under a common deployment oracle. Together these analyses establish portfolio construction as an audited axis without identifying a common causal mechanism.

We therefore conclude that routeability headroom is necessary but not sufficient for learnable LLM routing. Portfolio construction is an audited axis, not the sole causal mechanism: the data do not isolate portfolio-induced diversity from score granularity or tied-best saturation as ultimate causes. The declared oracle and comparator remain part of every portfolio claim.

\begin{figure}[t]
  \centering
  \includegraphics[width=\linewidth]{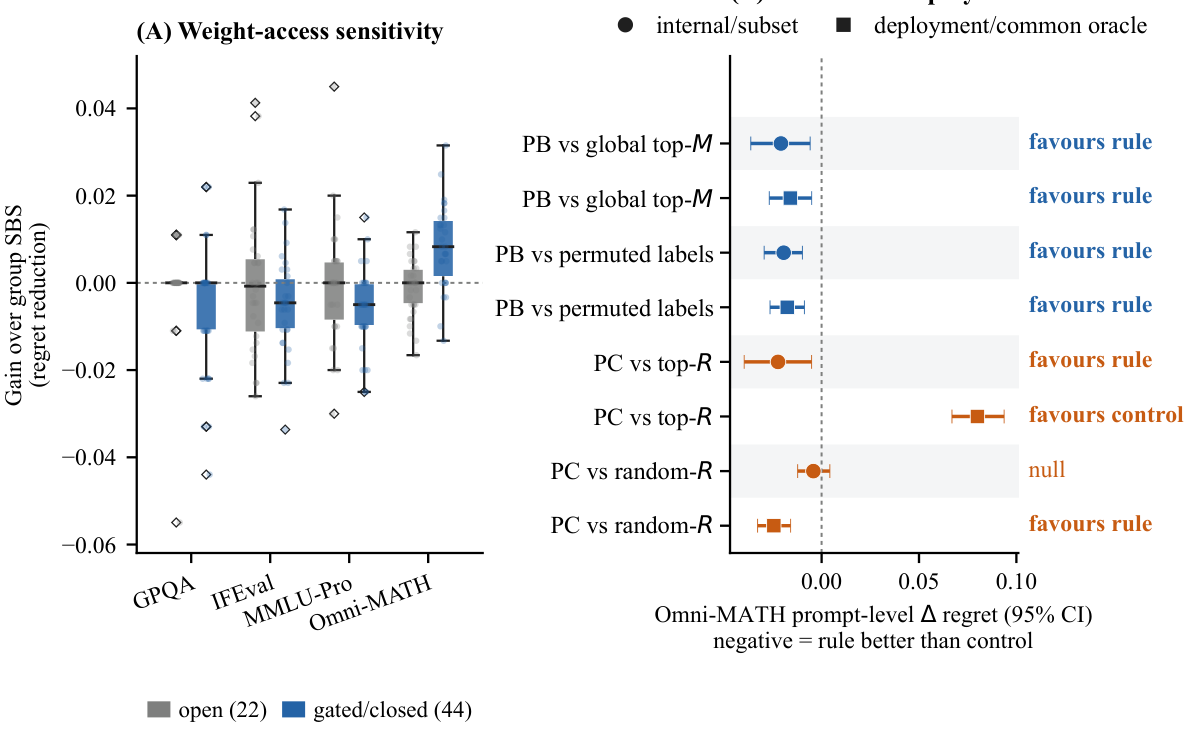}
  \caption{Portfolio evidence under declared estimands. \textbf{(A)} Corrected 22/44 weight-access sensitivity; boxplots and overlaid points summarize all 30 repeated split-level gains per group and are descriptive because splits overlap. The registered 40-test split-level Wilcoxon--Holm value is $0.004103$ ($0.005356$ when four near-zero differences are treated as zeros; Table~\ref{tab:app-access-group}). \textbf{(B)} Prompt-space comparisons survive both the internal and full-test controls, whereas the LLM-side comparison reverses under the common full-portfolio oracle. Circles indicate internal/subset comparisons; squares indicate full-test/common-oracle deployment comparisons.}
  \label{fig:convergence}
\end{figure}

\section{Parameter-Size Sensitivity of the Cost--Quality Trade-off}
\label{sec:cost}

Our cost audit uses a parameter-size proxy rather than real provider API prices. Under this proxy, gains are small and concentrated: Omni-MATH reaches hypervolume-vs-SBS $0.004255$ (bootstrap interval $[0.00146,0.00717]$), while GPQA, IFEval, and MMLU-Pro reach $0.000196$, $0.001485$, and $0.000343$. The conservative RouteLLM-A anchor has AUCs of $0.52$, $0.67$, $0.56$, and $0.59$.
Because no version-matched prices or controlled latency measurements are available, this is a parameter-size sensitivity analysis; deployment-cost evaluation remains future work.

\section{Robustness, Scope, and Reproducibility}
\label{sec:robustness}

The audits above change several conclusions relative to a seed-level reading. We therefore keep the corrected claim status, dependence sensitivity, and artifact checks in the main paper.

\subsection{What survives the corrected inference units?}

Table~\ref{tab:main-claim-audit} separates router-vs-router, router-vs-SBS, and common-oracle questions. Selective routing differs from forced routing in the prespecified prompt-level directions, but corrected Omni-MATH forced-vs-SBS does not cross $0.05$. PCA-64 is withdrawn; IFEval ExtraTrees remains favorable to Ridge without establishing SBS dominance. The LLM-side $K=8$ result reverses between its restricted and full-portfolio oracles.

\begin{table}[t]
  \caption{Corrected claim audit. Intervals are ordinary prompt-cluster bootstrap 95\% intervals and are not multiplicity-adjusted. Holm values are computed within the family named in Table~\ref{tab:stat-families} (PB/PC rows: ten-test original-score family; Appendix~\ref{app:prereg-sensitivity} gives the 30-test family and binarized scores). NB denotes the Nadeau--Bengio overlapping-split sensitivity; no headline row survives NB family correction.}
  \label{tab:main-claim-audit}
  \centering
  \scriptsize
  \resizebox{\linewidth}{!}{%
  \begin{tabular}{lllll}
    \toprule
    Comparison & $\Delta$ & 95\% interval & prompt Holm & Decision \\
    \midrule
    Omni forced vs SBS & $-0.0136$ & $[-0.0251,-0.0020]$ & $0.0912$ & downgraded \\
    MMLU PCA-64 vs no PCA & $-0.0017$ & $[-0.0051,+0.0018]$ & not retained & withdrawn \\
    IFEval ExtraTrees vs Ridge & $-0.0164$ & $[-0.0266,-0.0063]$ & $0.0126$ & retained, predictor-only \\
    IFEval RandomForest vs Ridge & --- & $[-0.0233,+0.0002]$ & not retained & unstable \\
    Omni prompt clusters vs global & $-0.0160$ & $[-0.0269,-0.0051]$ & $0.0140$ & protocol-specific \\
    Omni prompt clusters vs permuted & $-0.0176$ & $[-0.0265,-0.0088]$ & $0.0015$ & protocol-specific \\
    Omni LLM medoids vs top-$R$, own oracle & $-0.0224$ & $[-0.0397,-0.0051]$ & $0.0256$ & internal gap \\
    Omni LLM medoids vs top-$R$, full oracle & $+0.0800$ & $[+0.0669,+0.0937]$ & $0.0010$ & reverses \\
    \bottomrule
  \end{tabular}}
\end{table}

The 30 splits reuse prompts and are correlated. A follow-up plan frozen before real-data re-analysis specified prompt-level inference plus a Nadeau--Bengio corrected-$t$ sensitivity; the original submission itself was not preregistered. The smallest within-family NB-adjusted value is $0.212$. We retain prompt-level comparisons and do not read seed-level stability as independent replication.

\subsection{External range check and bounded generalization}

Applied post hoc to all 12 public RouterEval tasks~\citep{huang2025routereval}, RAP spans 55--1430 test prompts and 3--1000 candidate models. Ridge is favorable to task-specific SBS on 9/12 tasks (median delta $-0.0077$), and kNN-5 on 3/12; the exact sign test across the two predictors is not Holm-significant ($p_{\mathrm{Holm}}=0.146$). This confirms executability over a wider size range, not external validation or a universal router.

The completed utility sweep covers $K=7,13,33,50,66$; $K=20,59$ are structural-only. Provider-stratified draws and leave-one-out deletion are locally stable, but two whole-provider removals change regime labels. An updated portfolio is a new snapshot (benchmark version, model list, score schema, split manifest), not a free robustness replicate.

Numerical verification results for frozen inputs, splits, alignment, and claim-level statistical checks are summarized in Appendix~\ref{app:dataset} (Table~\ref{tab:main-repro-audit}).

\section{Limitations and Conclusion}
\label{sec:limitations}

\paragraph{Limitations.}
The empirical scope is four single-turn benchmarks; multi-turn, tool-use, agentic, preference, and deployment-cost settings are outside it. Conclusions are snapshot-conditional, and cost is a parameter-size proxy. No headline row survives the NB family sensitivity (Appendix~\ref{app:prereg-sensitivity}). ID checks do not establish absence of text overlap or proprietary-model pretraining contamination. Legacy no-chunk exports with non-finite values are disclosed, rejected by audited loaders, and excluded from release claims.

\paragraph{Conclusion.}
LLM routing is routeability-dependent: Omni-MATH contains retained but protocol-specific positive cells, while IFEval remains predictor-sensitive. The Routeability Audit Protocol (RAP) and RouteabilityBench provide a retrospective framework for auditing routing claims before treating an observed gain as general evidence. A prospective, partial-evaluation routeability diagnostic is a possible direction for future work, not a capability established here.

\paragraph{Competing interests.}
The authors declare no competing interests.

\section*{Acknowledgments}
\begingroup\footnotesize
Y.~Lu acknowledges support from the European Union's Horizon Europe research and innovation programme under the Marie Sk\l odowska-Curie Actions (MSCA), Grant Agreement No.~101081674 (SOUND.AI). D.~Kjorvezir, A.~Gjorgjevikj, and T.~Eftimov acknowledge funding from the Slovenian Research and Innovation Agency (ARIS) under research programme P2-0098 and research projects J2-70078 and GC-0001, as well as from the European Union under Grant Agreements No.~101211695 (Horizon Europe MSCA-PF AutoLLMSelect) and No.~101187010 (Horizon Europe ERA Chair AutoLearn-SI). C.~Doerr acknowledges funding from the European Research Council (ERC) under the European Union's Horizon Europe research and innovation programme (Grant Agreement No.~101125586, dynaBBO). This research was also jointly funded by the French National Research Agency (ANR-23-CE23-0035) and the German Research Foundation (DFG, LI 2801/7-1) through the Opt4DAC project. Views and opinions expressed are, however, those of the authors only and do not necessarily reflect those of the European Union or the European Research Council Executive Agency. Neither the European Union nor the granting authorities can be held responsible for them.
\endgroup

\bibliographystyle{plainnat}
\bibliography{references}

\appendix
\linespread{0.95}\selectfont
\setlength{\textfloatsep}{7pt plus 2pt minus 2pt}
\setlength{\intextsep}{7pt plus 2pt minus 2pt}
\renewcommand{\thetable}{A.\arabic{table}}
\setcounter{table}{0}
\renewcommand{\theHtable}{appendix.\arabic{table}}
\renewcommand{\thefigure}{A.\arabic{figure}}
\setcounter{figure}{0}
\renewcommand{\theHfigure}{appendix.\arabic{figure}}

\section{Statistical Tables}
\label{app:stats}

The appendix is organized as an evidence warehouse. Main numerical claims are indexed by hypothesis-family IDs, raw audit footprints, verification status, and short pseudocode. The paper reports evidence-family identifiers and reader-facing summaries rather than implementation-level bookkeeping.

Original seed-level screens used paired Wilcoxon signed-rank tests with within-family Holm correction. Corrected primary claims use prompt-cluster bootstrap intervals and prompt-level sign-flip tests with separately registered Holm families; overlapping-split Nadeau--Bengio analysis is a sensitivity check. The appendix-level statistical reporting follows standard classifier-comparison and reproducibility guidance, including classical rank-based tests, paired comparison cautions, NLP significance guidance, and benchmark-variance reporting~\citep{friedman1937ranks, demsar2006statistical, dietterich1998approximate, benavoli2017time, dror2018hitchhiker, bouthillier2021accounting, pineau2021reproducibility}. The routed action uses the same tie-aware rule as the main text, and Table~\ref{tab:app-stat-artifacts} records the audit footprint and verification status for each family.

\begin{table}[htbp]
  \caption{Hypothesis-family summary. Negative mean differences indicate lower regret for the left method where the underlying metric is absolute routed regret.}
  \label{tab:app-stat-families}
  \centering
  \scriptsize
  \setlength{\tabcolsep}{3pt}
  \begin{tabular}{>{\raggedright\arraybackslash}p{0.06\linewidth} >{\raggedright\arraybackslash}p{0.22\linewidth} >{\raggedright\arraybackslash}p{0.08\linewidth} >{\raggedright\arraybackslash}p{0.15\linewidth} >{\raggedright\arraybackslash}p{0.12\linewidth} >{\raggedright\arraybackslash}p{0.28\linewidth}}
    \toprule
    ID & Family & \# tests & Metric & Alternative & Key result \\
    \midrule
    H1 & Full/selective routing & 40 & gap closure / regret & left $<$ right & Omni-MATH forced vs risk-controlled, $p_{\text{Holm}}=1.57{\times}10^{-4}$ \\
    H2 & SBS dominance & 12 plus re-analysis & regret vs \sbs{} & prespecified & Omni-MATH forced downgraded; prompt Holm $p=0.0912$ \\
    H3 & Weight-access sensitivity & 40 & regret within group & method $<$ group \sbs{} & one taxonomy-sensitive Omni-MATH cell, $p_{\text{Holm}}=0.004103$ \\
    H4 & Portfolio estimands & 10 prompt-level & regret & declared per row & prompt-space full-test survives; LLM-side reverses under common oracle \\
    H5 & PCA audit & 24 plus re-analysis & forced regret & two-sided & PCA-64 withdrawn; 95\% interval includes zero \\
    H6 & Regressor family & 12 plus re-analysis & forced regret & left $<$ Ridge & ExtraTrees on IFEval survives, sign-flip Holm $p=0.0126$ \\
    H7 & RF sensitivity & 4 plus checks & forced regret & RF $<$ Ridge & unstable; no retained positive \\
    \bottomrule
  \end{tabular}
\end{table}

\begin{table}[t]
  \caption{Evidence artifact registry.}
  \label{tab:app-stat-artifacts}
  \centering
  \scriptsize
  \setlength{\tabcolsep}{3pt}
  \begin{tabular}{>{\raggedright\arraybackslash}p{0.06\linewidth} >{\raggedright\arraybackslash}p{0.23\linewidth} >{\raggedright\arraybackslash}p{0.22\linewidth} >{\raggedright\arraybackslash}p{0.20\linewidth} >{\raggedright\arraybackslash}p{0.19\linewidth}}
    \toprule
    ID & Evidence artifact & Raw footprint & Verification status & Role in paper \\
    \midrule
    H1 & Full/selective routing family & 40-test family; 30 seeds & frozen; Holm family checked & selective behavior is regime-dependent \\
    H2 & \sbs{} dominance audit & 12 screens plus prompt re-analysis & aligned prompt domain checked & positive claim downgraded \\
    H3 & Weight-access audit & 40-test family; 30 seeds & 66 taxonomy rows reviewed & taxonomy-sensitive cell \\
    H4 & LLM-side clustering audit & 48-test family; 9000 raw rows & frozen; H1/H2 caveat checked & top-$R$ homogeneity-trap evidence \\
    H5 & PCA audit & 24 screens plus prompt re-analysis & train-only PCA verified & PCA-64 withdrawn \\
    H6 & Regressor-family audit & 12-test family; 480 raw rows; 0 failed cells & frozen; diagnostics checked & ExtraTrees improves IFEval \\
    H7 & RandomForest sensitivity & 4 screens plus environment checks & realizations disagree & unstable; no positive claim \\
    \bottomrule
  \end{tabular}
\end{table}

\begin{quote}\footnotesize\ttfamily
Algorithm A1: Paired Wilcoxon + Holm over a named family.\\
Input: family $\mathcal{F}$ of hypotheses, each with 30 paired seed-level metric values.\\
For each $h \in \mathcal{F}$, compute paired differences $d^{(h)}_s = m^{(h)}_{\text{left},s} - m^{(h)}_{\text{right},s}$.\\
Run a paired Wilcoxon signed-rank test on $\{d^{(h)}_s\}_{s=1}^{30}$ using the direction declared for that family.\\
Collect the raw $p$-values within $\mathcal{F}$ and apply Holm correction only within that named family.\\
Report family size, corrected $p$-values, effect direction, and whether the Holm-adjusted decision is significant at $\alpha=0.05$.
\end{quote}

\subsection{Routeability mechanism diagnostics}

\begin{table}[h]
  \caption{Routeability mechanism diagnostics. Mechanism diagnostics characterize the routeability substrate; they are not used as a fitted predictive model.}
  \label{tab:app-mechanism-diagnostics}
  \centering
  \scriptsize
  \begin{tabular}{lrrrrrl}
    \toprule
    Dataset & Headroom & Forced GC & Score levels & Median tied-best & Unique-best & Regime label \\
    \midrule
    GPQA & 0.193 & -0.135 & 2 & 54.5\% & 1.1\% & high-headroom tied-best boundary \\
    IFEval & 0.045 & -0.324 & 5 & 87.9\% & 0.0\% & predictor-sensitive boundary \\
    MMLU-Pro & 0.091 & -0.147 & 2 & 75.8\% & 0.8\% & tied-best / low-utility boundary \\
    Omni-MATH & 0.200 & +0.081 & 5 & 36.4\% & 8.4\% & portfolio-sensitive positive regime \\
    \bottomrule
  \end{tabular}
\end{table}

Table~\ref{tab:app-mechanism-diagnostics} is the submission-facing mechanism summary. It is used to characterize score granularity and tied-best mass, not to fit or validate a forecasting rule.

\begin{table*}[t]
  \caption{Headline claim uncertainty summary. Prompt-level intervals are cluster bootstraps; ``seed agg.'' marks a seed-level sensitivity statistic. Negative deltas favor the left method for absolute regret. Sign-flip $p$-values use 10,000 permutations. $\Delta$ is the mean over bootstrap resamples of prompts; the observed seed-averaged difference differs by at most $0.003$ (e.g., $-0.0142$ for H2-SBS). Appendix~\ref{app:prereg-sensitivity} reports split-dependence and family-membership sensitivity.}
  \label{tab:app-claim-uncertainty}
  \centering
  \tiny
  \setlength{\tabcolsep}{2.5pt}
  \resizebox{\textwidth}{!}{%
  \begin{tabular}{>{\raggedright\arraybackslash}p{0.07\linewidth} >{\raggedright\arraybackslash}p{0.17\linewidth} >{\raggedleft\arraybackslash}p{0.07\linewidth} >{\raggedright\arraybackslash}p{0.14\linewidth} >{\raggedleft\arraybackslash}p{0.07\linewidth} >{\raggedleft\arraybackslash}p{0.07\linewidth} >{\raggedleft\arraybackslash}p{0.06\linewidth} >{\raggedleft\arraybackslash}p{0.04\linewidth} >{\raggedright\arraybackslash}p{0.15\linewidth}}
    \toprule
    claim & dataset / protocol & delta & CI & $p$ & $p_H$ & effect & $n$ & interpretation \\
    \midrule
    H1-selective & Omni-MATH forced vs risk-controlled & $-0.0158$ & $[-.0247,-.0069]$ & -- & $0.0040$ & $-0.913$ & 998 & prompt direction survives \\
    H2-SBS & Omni-MATH forced vs SBS & $-0.0136$ & $[-.0251,-.0020]$ & $0.0228$ & $0.0912$ & $-0.710$ & 998 & family claim downgraded \\
    H3-access & Omni-MATH gated-or-closed margin abstention & $-0.0085$ & seed agg. & $0.000103$ & $0.0041$ & $-0.818$ & 30 & taxonomy-sensitive \\
    H4-PB & full-test vs global top-$M$ & $-0.0160$ & $[-.0269,-.0051]$ & -- & $0.0140$ & $-0.888$ & 996 & deployment estimand survives \\
    H4-PB & full-test vs permuted labels & $-0.0176$ & $[-.0265,-.0088]$ & -- & $0.0015$ & $-0.927$ & 996 & deployment estimand survives \\
    H4-PC & subset-oracle vs top-$R$ & $-0.0224$ & $[-.0397,-.0051]$ & -- & $0.0256$ & $-0.703$ & 998 & internal gap closure \\
    H4-PC & full-oracle vs top-$R$ & $+0.0800$ & $[.0669,.0937]$ & -- & $0.0010$ & $0.974$ & 998 & interpretation reverses \\
    H5-PCA & MMLU-Pro PCA-64 vs raw Ridge & $-0.0017$ & $[-.0051,.0018]$ & -- & $0.3272$ & $-0.274$ & 999 & withdrawn \\
    H6-ET & IFEval ExtraTrees vs Ridge & $-0.0164$ & $[-.0266,-.0063]$ & -- & $0.0126$ & $-0.776$ & 540 & survives \\
    H7-RF & IFEval RandomForest vs Ridge & $-0.0117$ & $[-.0233,.0002]$ & -- & $0.1446$ & $-0.641$ & 540 & unstable; no positive \\
    \bottomrule
  \end{tabular}}
\end{table*}

Appendix claim-to-artifact mapping is documented in the supplementary release notes, which pair each headline claim with its frozen evidence table and the documented reproduction commands.

\section{Frozen-Plan Split-Dependence Sensitivity}
\label{app:prereg-sensitivity}

The follow-up plan was frozen on 2026-07-12 before real-data output from the re-analysis families was inspected. It specified prompt-level inference plus a Nadeau--Bengio corrected $t$ sensitivity for overlapping splits. The original submission itself was not pre-registered. Because the 30 splits reuse prompts, treating split means as independent replicates can be anticonservative. No headline row survives within-family Holm correction of the Nadeau--Bengio values; the minimum corrected value is $0.212$ (the adverse common-oracle LLM-side row; $0.481$ for the smallest favorable row of Table~\ref{tab:app-nb}). Primary positive statements are therefore prompt-level results that this split-level dependence approximation does not independently confirm.

\begin{table}[htbp]
  \caption{Primary prompt-level Holm values and split-dependence sensitivity. PB denotes prompt-space conditioning and PC LLM-side representatives. The $0.212$ minimum belongs to the adverse PC full-oracle row.}
  \label{tab:app-nb}
  \centering
  \scriptsize
  \begin{tabular}{lrr}
    \toprule
    Comparison & primary $p_{\rm Holm}$ & NB $p_{\rm Holm}$ \\
    \midrule
    Omni-MATH forced vs risk-controlled & $0.0040$ & $0.664$ \\
    Omni-MATH forced vs SBS & $0.0912$ & $1.000$ \\
    IFEval ExtraTrees vs Ridge & $0.0126$ & $1.000$ \\
    PB full-test vs global top-$M$ & $0.0140$ & $0.481$ \\
    PB full-test vs permuted labels & $0.0015$ & $0.481$ \\
    PC own-oracle vs top-$R$ & $0.0256$ & $1.000$ \\
    PC full-oracle vs top-$R$ & $0.0010$ & $0.212$ \\
    \bottomrule
  \end{tabular}
\end{table}

\paragraph{Score-version and family sensitivity of the portfolio rows.}
The frozen plan also specified two binarized versions of the Omni-MATH scores (correct only at $1.0$; correct at $\ge0.5$) next to the original five-level scores. Table~\ref{tab:app-binarized} repeats the portfolio estimands under all three. Holm values are given within the ten-test original-score family used in the main text ($p_{10}$) and within the 30-test family that pools the three score versions ($p_{30}$). The global-top-$M$ contrast depends on score granularity: it is null under strict binarization and favorable but not Holm-significant at the $0.5$ threshold. The contrasts against permuted labels keep their sign in all three versions, and the common-oracle LLM-side reversal appears in all three.

\begin{table}[htbp]
  \caption{Omni-MATH portfolio estimands under three score versions. $\Delta$ is the prompt-level bootstrap mean (95\% interval); negative favors the rule. $p_{10}$: Holm within the ten-test original-score family; $p_{30}$: Holm within the 30-test family pooling original, binarized-$1.0$ and binarized-$0.5$ rows.}
  \label{tab:app-binarized}
  \centering
  \tiny
  \setlength{\tabcolsep}{2.5pt}
  \resizebox{\linewidth}{!}{%
  \begin{tabular}{lrrrrrrr}
    \toprule
    Contrast & orig.\ $\Delta$ & $p_{10}$ & $p_{30}$ & bin-1.0 $\Delta$ [interval] & $p_{30}$ & bin-0.5 $\Delta$ [interval] & $p_{30}$ \\
    \midrule
    PB full-test vs global top-$M$ & $-0.0160$ & $0.0140$ & $0.0420$ & $+0.0010\ [-0.0127,+0.0150]$ & $1.000$ & $-0.0155\ [-0.0279,-0.0027]$ & $0.135$ \\
    PB full-test vs permuted labels & $-0.0176$ & $0.0015$ & $0.0051$ & $-0.0163\ [-0.0253,-0.0072]$ & $0.0096$ & $-0.0158\ [-0.0252,-0.0063]$ & $0.0270$ \\
    PB internal vs global top-$M$ & $-0.0209$ & $0.0237$ & $0.0869$ & $+0.0020\ [-0.0147,+0.0189]$ & $1.000$ & $-0.0143\ [-0.0309,+0.0024]$ & $0.453$ \\
    PC own-oracle vs top-$R$ & $-0.0224$ & $0.0256$ & $0.1152$ & $-0.0292\ [-0.0481,-0.0111]$ & $0.0280$ & $-0.0256\ [-0.0447,-0.0071]$ & $0.0910$ \\
    PC full-oracle vs top-$R$ & $+0.0800$ & $0.0010$ & $0.0030$ & $+0.0596\ [+0.0459,+0.0741]$ & $0.0030$ & $+0.0744\ [+0.0602,+0.0888]$ & $0.0030$ \\
    \bottomrule
  \end{tabular}}
\end{table}

\paragraph{Reconciliation of the earlier H4/H5 screens.} The rows named H4 and H5 in \texttt{e1\_inference\_robustness.csv} are not repeated measurements of the PB/PC rows above. The earlier screen fits a train-only KMeans prompt partition and a Ridge router on each split over the 17 compact embeddings, and reports $+0.002581$ versus global top-$M$, $+0.003669$ versus permuted labels, and $+0.042451$ for centroid versus top-$R$ on 998 Omni-MATH prompts (positive is adverse under its declared regret convention). The PB/PC reconciliation instead uses frozen split assignments with hard train-mean mappings, declared same-subset or full-test prompt domains, and either a sub-portfolio or full-portfolio oracle; its full-test PB rows use 996 prompts and its PC rows use 998. These protocol and estimand differences explain the opposite directions, but do not turn them into independent confirmations. We retain the earlier adverse screen in the released evidence and interpret the PB/PC rows only as protocol-specific results.

\paragraph{Family membership of the prompt-level headline values.}
The nine-test headline family used for H2, H5, H6 and H7 contains one row built on the superseded 38/28 access taxonomy and four portfolio rows that the follow-up amendment replaced by the ten-test family above. Removing the taxonomy row leaves the four Holm values unchanged ($0.0912$, $0.3272$, $0.0126$, $0.1446$). Removing all five rows gives $0.0684$ (H2), $0.3272$ (H5), $0.0084$ (H6) and $0.0964$ (H7); no decision changes.

\paragraph{Inference-status comparison.}
The public rebuttal reported seed-level Wilcoxon--Holm values over 30 overlapping splits. After the frozen prompt-level plan and the Nadeau--Bengio sensitivity, the statuses in Table~\ref{tab:app-rebuttal} changed; the other quantities quoted in the rebuttal are unchanged. Each listed change moves the claim toward weaker evidence. Table~\ref{tab:app-binarized} additionally reports a score-version check that the rebuttal did not include.

\begin{table}[htbp]
  \caption{Statuses that differ between the author rebuttal (28 July) and this version.}
  \label{tab:app-rebuttal}
  \centering
  \scriptsize
  \begin{tabular}{>{\raggedright\arraybackslash}p{0.19\linewidth}>{\raggedright\arraybackslash}p{0.21\linewidth}>{\raggedright\arraybackslash}p{0.26\linewidth}>{\raggedright\arraybackslash}p{0.19\linewidth}}
    \toprule
    Item & Rebuttal & This version & Reason \\
    \midrule
    IFEval ExtraTrees vs Ridge & $p_{\rm Holm}=0.0019$ (seed-level, 12 tests) & $0.0126$ (sign-flip); predictor-only & prompt-level inference \\
    IFEval RandomForest vs Ridge & $0.0012$ (seed-level, 4 tests) & not retained; unstable & interval reaches zero; software realizations disagree \\
    MMLU-Pro PCA-64 & $0.019$ (seed-level, 24 tests) & withdrawn & $\Delta=-0.0017$, interval includes zero \\
    Prompt clusters vs global top-$M$ / permuted & $4.4{\times}10^{-4}$ / $1.1{\times}10^{-4}$ (seed-level, 56 tests) & $0.0140$ / $0.0015$ (prompt-level, 10 tests) & inference unit \\
    LLM-side $K=8$ vs top-$R$ & $0.012$ (seed-level, 48 tests) & own-oracle $0.0256$; common-oracle reversal $+0.0800$ & dual-estimand audit \\
    Routed-utility portfolio sizes & seven sizes & five ($K=20,59$ structural-only) & routed masking not run at $K=20,59$ \\
    \bottomrule
  \end{tabular}
\end{table}

\section{Reporting Checklist}
\label{app:reporting-checklist}

Table~\ref{tab:reporting} summarizes the compact reporting checklist derived from RAP and used by RouteabilityBench.

\begin{table}[h]
  \caption{Compact reporting checklist for LLM routing benchmarks.}
  \label{tab:reporting}
  \centering
  \small
  \begin{tabular}{p{0.30\linewidth} p{0.60\linewidth}}
    \toprule
    Item & Minimum reporting standard \\
    \midrule
    Routeability & Report SBS, VBS, headroom, tied-best distribution, and score granularity before routing gains. \\
    Portfolio & Disclose model list, provider-family counts, access groups, portfolio restrictions, and whether comparisons are within-group or full-portfolio. \\
    Predictors & Audit predictor family and representation choices; do not attribute all effects to the routing algorithm alone. \\
    Statistics & Use frozen multi-seed splits, paired tests, and Holm correction over named hypothesis families. \\
    Selective routing & Report calibration data, threshold protocol, coverage, and forced-vs-selective results separately. \\
    Cost & State whether costs are real prices, HELM-derived, GPU estimates, or proxies. \\
    Comparators/artifacts & Report constrained comparators only as limitations; provide data cards or equivalent artifact documentation. \\
    \bottomrule
  \end{tabular}
\end{table}

\section{Dataset Selection and Embedding Generation}
\label{app:dataset-embedding}

\subsection{Proof of Proposition 1}
\label{app:strict-winner-proof}

Conditioned on $\tieset$, the strict label $Z_i$ is uniform over $\tieset$. If $\pi(i)\in \tieset$, then $Y_{i,\pi(i)}=\max_k Y_{i,k}$, so tie-aware regret is zero. Exactly one of the $|\tieset|$ possible strict labels equals $\pi(i)$, so $\mathbb{P}(\pi(i)=Z_i\mid \tieset)=1/|\tieset|$. Therefore the expected strict-label accuracy is $1/|\tieset|$ even for a zero-regret policy.

\paragraph{Score-matrix provenance.}
The four frozen prompt-level score matrices, each with 66 model columns, are reconstructed cell by cell from public HELM Capabilities per-instance outputs~\citep{xu2025helmcapabilities}. Release v1.15.0 is the first index covering all 66 portfolio models. The headline metric is \texttt{chain\_of\_thought\_correctness} for GPQA and MMLU-Pro, \texttt{ifeval\_strict\_accuracy} for IFEval, and \texttt{omni\_math\_accuracy} for Omni-MATH. The frozen audit compares 197,142 cells with zero mismatches. Release v1.15.0 contains two additional models covering all four scenarios, so the audited portfolio is a fixed 66-model subset of 68 eligible models and all claims remain snapshot-conditional.

The retained tasks are GPQA, IFEval, MMLU-Pro, and Omni-MATH~\citep{rein2024gpqa, zhou2023ifeval, wang2024mmlupro, gao2025omnimath}; MMLU-Pro is interpreted relative to MMLU~\citep{hendrycks2021mmlu}. BIG-bench, AlpacaFarm, LiveBench, and benchmark-audit work provide broader context~\citep{srivastava2023bigbench, dubois2023alpacafarm, white2025livebench, raji2021benchmark, schaeffer2023mirage}. WildBench was excluded because its multi-turn structure changes the representation target.

\begin{table}[h]
  \caption{Prompt-domain alignment. Raw counts come from score matrices; aligned counts are the exact intersection with all three compact encoder views used for fresh refits.}
  \label{tab:app-prompt-length}
  \centering
  \small
  \begin{tabular}{lrrl}
    \toprule
    Dataset & Raw $N$ & Aligned $N$ & Excluded prompt IDs \\
    \midrule
    GPQA & 446 & 445 & \texttt{id344} \\
    IFEval & 541 & 541 & none \\
    MMLU-Pro & 1000 & 1000 & none \\
    Omni-MATH & 1000 & 998 & \texttt{id189}, \texttt{id4160} \\
    All retained & 2987 & 2984 & 3 rows \\
    \bottomrule
  \end{tabular}
\end{table}

\begin{table}[h]
  \caption{Embedding models and view configurations. Embedding views are audit inputs rather than a claimed routing method.}
  \label{tab:app-embedding-views}
  \centering
  \small
  \begin{tabular}{p{0.23\linewidth} p{0.34\linewidth} p{0.33\linewidth}}
    \toprule
    Component & Models / views & Details \\
    \midrule
    Encoder-style models & BGE, E5, GIST & dims $1024/1024/384$; approx.\ 512-token context \\
    Decoder embedding models & Qwen3-8B, Qwen3-4B, Nemotron-8B & dims $4096/2560/4096$; 4096-token no-chunk limit \\
    View configurations & compact encoder family; large no-chunk decoder family; large chunked decoder family & three-view families \\
    View configurations & mixed six-view chunked family; mixed six-view no-chunk family; full nine-view family & six-/nine-view families \\
    \bottomrule
  \end{tabular}
\end{table}

Pooling uses attention-mask-weighted mean pooling followed by L2 normalization. Compact encoders use non-overlapping 510-token chunks. The large no-chunk configuration may truncate beyond 4096 tokens; the chunked configuration uses overlapping 510-token windows with stride 256. Length weighting gives overlap tokens repeated weight, so it is an approximation. The representation audit is informed by MTEB, BEIR, Sentence-BERT, and the published E5, GISTEmbed, and BGE/C-Pack sources~\citep{muennighoff2023mteb, thakur2021beir, reimers2019sentencebert, wang2022e5, solatorio2024gistembed, xiao2024cpack, wang2024improvingembeddings}. The compact raw files contain the three null prompt rows in Table~\ref{tab:app-prompt-length}; the declared intersection is finite. Separate legacy \texttt{fixed\_nochunk} exports contain non-finite rows and are excluded from release claims; audited loaders use the finite export path and reject or drop non-finite rows.

\begin{quote}\footnotesize\ttfamily
Embedding generation pseudocode.\\
1. Tokenize the prompt text.\\
2. If the selected view is chunked, split the token sequence into windows; otherwise keep the full prompt.\\
3. Encode each chunk with the chosen embedding model.\\
4. Apply attention-mask-weighted mean pooling over final hidden states.\\
5. L2-normalize each pooled chunk vector.\\
6. Aggregate chunk vectors with length weights to form one prompt embedding.\\
7. Reject or log any null, NaN, or zero vector before writing the frozen matrix.
\end{quote}

\section{Full-Portfolio Clustering Diagnostics}
\label{app:tas008}

This section audits the original full-portfolio prompt-clustering idea before the later portfolio controls. Under the naive full-66 mapping, strict selector accuracy is at most $0.013$, the maximum configuration-level mean embedding-to-oracle ARI is $0.058$, and the routable-cluster fraction is approximately zero. This is a boundary result, not a claim that clustering is generally useless.

\begin{table}[h]
  \caption{Full-portfolio clustering diagnostics. These diagnostics identify a boundary case caused by tied-best saturation and weak alignment between prompt embedding geometry and LLM performance space.}
  \label{tab:app-full-cluster}
  \centering
  \small
  \begin{tabular}{ll}
    \toprule
    Diagnostic & Audited value / interpretation \\
    \midrule
    Strict selector accuracy & $\leq 0.013$ under the full-portfolio selector summary \\
    Embedding-to-oracle ARI & maximum configuration-level mean $0.058$ \\
    Routable-cluster fraction, $\alpha=0.05$ & $0$ across the decomposition table \\
    Routable-cluster fraction, $\alpha=0.10$ & approximately $0$; only a negligible GPQA cell appears \\
    Selective advantage over cluster SBS & maximum $0$ in the decomposition artifact \\
    \bottomrule
  \end{tabular}
\end{table}

The appendix therefore treats naive full-portfolio clustering as an ill-posed boundary case under coarse, tied-best-heavy scoring. The portfolio analyses in \S\ref{sec:portfolio} use different estimands and are reported separately.

\section{Margin Calibration and Coverage--Risk}
\label{app:task009a}

Selective and risk-controlled routing are useful stress tests because they separate confidence from decision utility. Prompt-level re-analysis confirms the four prespecified selective-versus-forced directions, but SBS dominance is stricter. The original seed-level operating-point screen treated overlapping splits as replicates and used a misaligned SBS prompt domain. After alignment and prompt-level re-analysis, Omni-MATH forced routing has $\Delta=-0.0136$ with interval $[-0.0251,-0.0020]$, but family-level sign-flip Holm $p=0.0912$; the SBS-dominance claim is downgraded.

\begin{table}[h]
  \caption{Selective-routing summary. Lower regret is better; gap closure is relative to SBS--VBS headroom and is not by itself an SBS-dominance claim.}
  \label{tab:app-selective}
  \centering
  \scriptsize
  \begin{tabular}{lccp{0.38\linewidth}}
    \toprule
    Dataset & Forced regret & Risk-controlled regret & SBS-dominance outcome \\
    \midrule
    GPQA & $0.219780$ & $0.205128$ & no operating point beats SBS after Holm \\
    IFEval & $0.062793$ & $0.053007$ & no operating point beats SBS after Holm \\
    MMLU-Pro & $0.109833$ & $0.099833$ & no operating point beats SBS after Holm \\
    Omni-MATH & $0.193671$ & $0.209204$ & favorable direction; prompt-level family claim downgraded \\
    \bottomrule
  \end{tabular}
\end{table}

\begin{table}[h]
  \caption{SBS-dominance summary. The first three rows retain legacy screen values; Omni-MATH reports the prompt-level re-analysis used for the camera-ready decision.}
  \label{tab:app-sbs-dominance}
  \centering
  \scriptsize
  \begin{tabular}{lccc}
    \toprule
    Dataset & Best operating point & $p_{\text{Holm}}$ & Beats \sbs{} \\
    \midrule
    GPQA & margin abstention & 0.331199 & no \\
    IFEval & risk-controlled & 0.731335 & no \\
    MMLU-Pro & risk-controlled & 0.731335 & no \\
    Omni-MATH & forced routing & 0.0912 & no (downgraded) \\
    \bottomrule
  \end{tabular}
\end{table}

\begin{table}[h]
  \caption{Corrected cross-dataset calibration transfer. Only the corrected comparisons are reported.}
  \label{tab:app-cross-dataset}
  \centering
  \scriptsize
  \begin{tabular}{lcc}
    \toprule
    Calibration $\rightarrow$ test & $p_{\text{Holm}}$ & Holm-significant \\
    \midrule
    Omni-MATH $\rightarrow$ GPQA & 1.000000 & no \\
    Omni-MATH $\rightarrow$ MMLU-Pro & 0.811668 & no \\
    Omni-MATH $\rightarrow$ IFEval & 0.350685 & no \\
    GPQA $\rightarrow$ Omni-MATH & 0.062439 & no \\
    MMLU-Pro $\rightarrow$ IFEval & 0.811668 & no \\
    GPQA $\rightarrow$ MMLU-Pro & 1.000000 & no \\
    \bottomrule
  \end{tabular}
\end{table}

Because all six corrected transfer rows are non-significant after Holm correction, with the smallest corrected value GPQA$\rightarrow$Omni-MATH at $p_{\text{Holm}}=0.062439$, we do not treat cross-dataset threshold transfer as a positive result. Instead, the appendix uses these rows only to support the narrower statement that calibration thresholds appear dataset-specific under the corrected protocol. This caution is consistent with the broader selective-prediction and calibration literature, where abstention, rejection, and confidence transfer must be interpreted through operating-point tradeoffs rather than a single generic win~\citep{elyaniv2010foundations, geifman2017selective, geifman2019selectivenet, cortes2016rejection, guo2017calibration}.

\section{RouteLLM Ensemble-of-3 Upper Bound}
\label{app:routellm}

This appendix adds a RouteLLM-style conservative anchor and an ensemble-of-3 appendix variant under the same parameter-size proxy used in \S\ref{sec:cost}. The classifier AUC means for the conservative RouteLLM-A baseline are GPQA $0.52$, IFEval $0.67$, MMLU-Pro $0.56$, and Omni-MATH $0.59$; the ensemble appendix variant uses the same classifier scores with a stronger-anchor cost construction. These rows are baseline context, not evidence of real provider-price dominance. They should be read alongside the broader cost-aware routing and inference-systems literature rather than as deployment-price claims~\citep{muqeeth2024phatgoose, kwon2023vllm, yu2022orca, pope2023scalinginference, zhong2024distserve}.

\begin{table}[h]
  \caption{RouteLLM-style conservative baseline under the parameter-size proxy. AUC summarizes the conservative binary routing classifier; costs are not real API prices.}
  \label{tab:app-routellm}
  \centering
  \small
  \begin{tabular}{lcc}
    \toprule
    Dataset & Conservative classifier AUC & Best proxy hypervolume vs SBS \\
    \midrule
    GPQA & $0.52$ & $0.000196$ \\
    IFEval & $0.67$ & $0.001485$ \\
    MMLU-Pro & $0.56$ & $0.000343$ \\
    Omni-MATH & $0.59$ & $0.004255$ \\
    \bottomrule
  \end{tabular}
\end{table}

\begin{table}[h]
  \caption{Cost-proxy audit for the 66-model portfolio. These rows describe provenance of the cost field rather than any deployment pricing claim.}
  \label{tab:app-cost-audit}
  \centering
  \scriptsize
  \begin{tabular}{lrrrr}
    \toprule
    Portfolio size & Usable rows & Provider-priced & HELM-derived & Parameter-size proxy \\
    \midrule
    66 & 66 & 0 & 0 & 66 \\
    \bottomrule
  \end{tabular}
\end{table}

The cost analysis should be read as a disclosure and stress test: all 66 portfolio models have usable parameter-size proxy rows, with no provider-priced, HELM-derived, or GPU-hour-estimated cost rows in the audited cost registry. All costs are parameter-size proxies, not real API prices.

\section{Regressor-Family Audit}
\label{app:regressor-family}

The four-regressor family contains $480$ raw rows: four datasets, 30 seeds, and four regressors (Ridge, PLS, kNN, and ExtraTrees), with no failed cells. Prompt-level re-analysis retains IFEval ExtraTrees versus Ridge ($\Delta=-0.0164$, 95\% interval $[-0.0266,-0.0063]$, sign-flip Holm $p=0.0126$). The other datasets show no corresponding retained gain.

A RandomForest realization sensitivity adds $120$ raw rows with no failed cells or duplicate keys. Its favorable seed-level IFEval screen does not survive the full robustness rule: the original prompt-level interval reaches zero while another software realization excludes zero. RandomForest is therefore marked computationally unstable rather than treated as evidence.

\begin{table}[h]
  \caption{Regressor-audit diagnostics and row counts.}
  \label{tab:app-regressor-diagnostics}
  \centering
  \scriptsize
  \begin{tabular}{lrrrr}
    \toprule
    Family & Raw rows & Duplicate keys & Failed cells & Holm family size \\
    \midrule
    Four-regressor family & 480 & 0 & 0 & 12 \\
    RandomForest realization sensitivity & 120 & 0 & 0 & 4 \\
    \bottomrule
  \end{tabular}
\end{table}

\begin{table}[h]
  \caption{Legacy seed-level regressor screen. These Holm values are sensitivity summaries; Table~\ref{tab:app-claim-uncertainty} gives the prompt-level decisions, under which ExtraTrees survives and RandomForest does not.}
  \label{tab:app-regressor}
  \centering
  \small
  \begin{tabular}{lrrrr}
    \toprule
    Dataset & ExtraTrees $\Delta$ & ExtraTrees $p_{\text{Holm}}$ & RandomForest $\Delta$ & RandomForest $p_{\text{Holm}}$ \\
    \midrule
    GPQA & $-0.0007$ & $1.0000$ & $-0.0040$ & $0.2369$ \\
    IFEval & $-0.0184$ & $0.0019$ & $-0.0178$ & $0.0012$ \\
    MMLU-Pro & $-0.0010$ & $1.0000$ & $-0.0043$ & $0.2369$ \\
    Omni-MATH & $-0.0062$ & $0.5522$ & $-0.0069$ & $0.2369$ \\
    \bottomrule
  \end{tabular}
\end{table}

IFEval is therefore predictor-sensitive. ExtraTrees improves on Ridge under the audited comparison, but the analysis does not show that either predictor beats the SBS.

\section{Portfolio-Construction Evidence}
\label{app:portfolio-warehouse}

This appendix expands the weight-access, prompt-space, and LLM-side portfolio analyses. They share the same score matrix and use different estimands, so they are reported separately rather than as independent converging evidence.

\begin{table*}[t]
  \caption{Corrected weight-access sensitivity for margin abstention. Positive gain means lower routed regret than the group \sbs{}. Adjusted values use the complete 40-test family. In the Omni-MATH gated-or-closed row, four of 30 seeds have $|$difference$|<10^{-8}$ (float32 noise). The primary $p_{\text{Holm}}$ ($0.0041$) ranks them as non-zero; treating them as zeros, as the reported effect size does, gives $0.0054$; exact enumeration gives $0.0014$. All are below $0.05$.}
  \label{tab:app-access-group}
  \centering
  \scriptsize
  \setlength{\tabcolsep}{3pt}
  \begin{tabular}{>{\raggedright\arraybackslash}p{0.10\linewidth} >{\raggedright\arraybackslash}p{0.16\linewidth} >{\raggedright\arraybackslash}p{0.26\linewidth} rrc}
    \toprule
    Dataset & Weight-access group & Method & Mean gain & $p_{\text{Holm}}$ & Significant \\
    \midrule
    GPQA & open-weight & margin abstention & $-0.000733$ & 1.000000 & no \\
    GPQA & gated-or-closed-weight & margin abstention & $-0.006960$ & 1.000000 & no \\
    IFEval & open-weight & margin abstention & $-0.000459$ & 1.000000 & no \\
    IFEval & gated-or-closed-weight & margin abstention & $-0.005708$ & 1.000000 & no \\
    MMLU-Pro & open-weight & margin abstention & $-0.000167$ & 1.000000 & no \\
    MMLU-Pro & gated-or-closed-weight & margin abstention & $-0.006500$ & 1.000000 & no \\
    Omni-MATH & open-weight & margin abstention & $-0.001382$ & 1.000000 & no \\
    Omni-MATH & gated-or-closed-weight & margin abstention & $+0.008458$ & 0.004103 & yes \\
    \bottomrule
  \end{tabular}
\end{table*}

\subsection{External size-range check on RouterEval}

We applied RAP post hoc to all 12 public RouterEval tasks~\citep{huang2025routereval}, spanning 55--1430 test prompts and candidate pools of 3--1000 models. Ridge is favorable relative to task-specific SBS on 9 of 12 tasks (median task delta $-0.0077$), but the exact sign test is not significant after Holm correction across Ridge and kNN-5 ($p_{\text{Holm}}=0.146$); kNN-5 is favorable on only 3 of 12. This is a bounded, predictor-dependent scope check, not external validation of the four submitted benchmark labels.

\begin{table}[htbp]
  \caption{Prompt-space control evidence after dual-estimand prompt-level re-analysis. Negative delta means lower regret for the cluster-conditioned rule.}
  \label{tab:app-prompt-cluster}
  \centering
  \scriptsize
  \setlength{\tabcolsep}{3pt}
  \begin{tabular}{>{\raggedright\arraybackslash}p{0.13\linewidth} >{\raggedright\arraybackslash}p{0.15\linewidth} >{\raggedright\arraybackslash}p{0.22\linewidth} rrc}
    \toprule
    Dataset & Estimand & Control & $\Delta$ & 95\% CI & $p_{\text{Holm}}$ \\
    \midrule
    Omni-MATH & same-subset & global top-$M$ & -0.0209 & $[-0.0364,-0.0059]$ & 0.0237 \\
    Omni-MATH & full-test & global top-$M$ & -0.0160 & $[-0.0269,-0.0051]$ & 0.0140 \\
    Omni-MATH & same-subset & permuted labels & -0.0195 & $[-0.0295,-0.0099]$ & 0.0012 \\
    Omni-MATH & full-test & permuted labels & -0.0176 & $[-0.0265,-0.0088]$ & 0.0015 \\
    \bottomrule
  \end{tabular}
\end{table}

\begin{table}[htbp]
  \caption{LLM-side clustering dual-estimand evidence. Negative regret difference favors centroid-medoid representatives.}
  \label{tab:app-llm-cluster}
  \centering
  \scriptsize
  \begin{tabular}{lcrrc}
    \toprule
    Dataset / $K$ & Estimand / control & $\Delta$ & 95\% CI & Interpretation \\
    \midrule
    Omni-MATH / 8 & subset oracle / top-$R$ & -0.0224 & $[-0.0397,-0.0051]$ & internal gap closure \\
    Omni-MATH / 8 & subset oracle / random-$R$ & -0.0042 & $[-0.0123,0.0042]$ & null \\
    Omni-MATH / 8 & full oracle / top-$R$ & +0.0800 & $[0.0669,0.0937]$ & top-$R$ better \\
    Omni-MATH / 8 & full oracle / random-$R$ & -0.0245 & $[-0.0328,-0.0159]$ & centroid better \\
    \bottomrule
  \end{tabular}
\end{table}

The supporting evidence for this section covers 66 reviewed taxonomy rows, corrected access-group summaries and raw rows (8 / 1440 / 40), prompt-cluster audit and robustness tables (6 / 1890 / 56 / 17640), and LLM-side clustering tables (48 / 12 / 9000). The dual-estimand rows distinguish internal gap closure from deployment-relevant achieved score; the protocols are not independent replications.

\section{Reproducibility and Artifact Manifest}
\label{app:dataset}

The RouteabilityBench artifact contains frozen splits, an authoritative filtered-row--prompt-ID registry, evidence tables, metadata, and executable claim-verification instructions. Its documentation follows datasheet, model-card, and data-card patterns~\citep{gebru2021datasheets, mitchell2019modelcards, pushkarna2022datacards}; Croissant applicability depends on the final legally approved release scope~\citep{akhtar2024croissant}. Raw prompts, model responses, embeddings, prediction caches, and HELM-derived score matrices are not redistributed. Instead, a pinned manifest and fetch script reconstruct the four matrices from the official HELM Capabilities per-instance outputs and verify their canonical hashes before any split is applied. The corrected SBS family then reproduces from those locally reconstructed matrices. Exact ties for the best training mean occur in one or two seeds per dataset and are resolved by floating-point comparison of the training means rather than by a stated rule; resolving them by first or last model index instead changes the $K=66$ row of Table~\ref{tab:portfolio-size} by at most $0.002$ and leaves its signs unchanged (the near-equal IFEval and MMLU-Pro entries can swap order). Router refitting is deliberately outside the minimal artifact because it would require non-distributed embeddings, prediction caches, and training code.

The supplementary artifact supplies the frozen tables and scripts needed for the claim classes marked ZIP-runnable in its claim-to-artifact ledger, including official-source HELM score reconstruction and hash verification. It does not support end-to-end embedding generation, router refitting, or every figure render. Those boundaries are part of the artifact contract rather than being inferred from missing files.

\begin{table}[htbp]
  \caption{Executed numerical verification checks.}
  \label{tab:main-repro-audit}
  \centering
  \small
  \begin{tabular}{p{0.38\linewidth}p{0.19\linewidth}p{0.32\linewidth}}
    \toprule
    Check & Result & Interpretation \\
    \midrule
    Frozen SHA-256 manifest & 28/28 match & inputs unchanged \\
    ZIP split/alignment audit & 120/120 pass & exact filtered-domain coverage; 2,984 ID rows \\
    ZIP-only pytest functions & 6/6 pass & artifact-file evidence checks only \\
    ZIP-only HELM reconstruction & 197,142/197,142 & official-source fetch; canonical hashes match \\
    \bottomrule
  \end{tabular}
\end{table}

\begin{table}[htbp]
  \caption{Compute-resources disclosure for reviewer-side reproduction. The submission artifact is designed around frozen evidence tables and local score reconstruction rather than rerunning raw LLM benchmark scoring.}
  \label{tab:app-compute}
  \centering
  \scriptsize
  \begin{tabular}{p{0.20\linewidth}p{0.24\linewidth}p{0.26\linewidth}p{0.13\linewidth}}
    \toprule
    Component & Compute setting & Reproduction scope & Approx. time \\
    \midrule
    Paper compilation & local workstation; macOS; TeX Live 2025 & compile manuscript and checklist from source & minutes \\
    Artifact claim checks & CPU-only; reconstructed score matrices, frozen splits, and evidence tables & regenerate Table 1/Table 3 summaries and registered statistical families; verify prompt alignment & minutes after download \\
    Router refitting & not included in the artifact & requires non-distributed embeddings, prediction caches, and training code & outside artifact scope \\
    Raw benchmark scoring & not included in the artifact & upstream LLM evaluation is outside the shipped verification path & not measured \\
    \bottomrule
  \end{tabular}
\end{table}

\begin{table}[htbp]
  \caption{Submission artifact components for the E\&D release package.}
  \label{tab:app-artifact-plan}
  \centering
  \small
  \begin{tabular}{p{0.27\linewidth} p{0.62\linewidth}}
    \toprule
    Component & Purpose \\
    \midrule
    Data card / metadata & Document retained benchmarks, prompt counts, score levels, upstream licenses, and pinned canonical hashes; no data are shipped. \\
    Benchmark provenance & Record source benchmark versions and preprocessing choices. \\
    Portfolio documentation & List all 66 models with weight-access, service-access, and provider metadata. \\
    Frozen splits & Provide train/validation/calibration/test partitions for all 30 seeds. \\
    Derived score/evidence tables & Provide paper-facing routing summaries, uncertainty tables, and audit outputs. \\
    Filtered-row registry & Map split indices to prompt IDs and source score rows, with fail-loud validation. \\
    Official-source reconstruction & Source manifest (run, suite, SHA-256) and program that rebuild the four score matrices from public HELM outputs. \\
    ID/content-hash utility & Standard-library tool to compare licensed local benchmark copies by exact ID and canonical content hash. \\
    Claim-verification scripts & Recreate the reported CSV summaries and registered statistical families from candidate files. \\
    Internal-only components & Score matrices (rebuilt by the reader), embeddings, prediction caches, router-training code, and figure-rendering code are not in this minimal artifact. \\
    Release cleanup & Complete anonymity, license, and redistribution checks before public release. \\
    \bottomrule
  \end{tabular}
\end{table}

\begin{table}[htbp]
  \caption{Supplementary artifact components.}
  \label{tab:app-repro-manifest}
  \centering
  \scriptsize
  \setlength{\tabcolsep}{3pt}
  \begin{tabular}{>{\raggedright\arraybackslash}p{0.27\linewidth} >{\raggedright\arraybackslash}p{0.33\linewidth} >{\raggedright\arraybackslash}p{0.28\linewidth}}
    \toprule
    Component & Contents & Verification role \\
    \midrule
    Frozen split manifest & 30-seed train/validation/calibration/test partitions for all retained benchmarks & fixes the data partitioning protocol \\
    Prompt-ID row registry & filtered index, source score row, and prompt ID for 2,984 valid rows & prevents silent split/score misalignment \\
    HELM rebuild manifest & 264 official files (run, suite, SHA-256), instance order, and canonical hashes of the four matrices & lets readers re-derive Table~1 and the alignment check without redistribution \\
    ID/content-hash utility & standard-library canonicalization and hashing of licensed local copies & exact ID/content comparison; no statement about pretraining contamination \\
    Derived score/evidence tables & family-level and seed-level statistical results, metadata, and uncertainty summaries & supports reported $p$-values, row counts, and significance calls \\
    Evidence tables & family-level inputs, sub-portfolio masking rows, access-group and selective-routing seed rows, view summary, taxonomy, and RouterEval summary & supports reported $p$-values, row counts, and significance calls \\
    Claim-verification code & table, family, split, and alignment checks & recomputes the supported numerical summaries; does not render every paper figure \\
    Reproducibility README & exact commands and explicit reproduction classes & separates ZIP-runnable checks from internal-only audits \\
    Release cleanup checklist & anonymity, licensing, and redistribution review items & records release status before public sharing \\
    \bottomrule
  \end{tabular}
\end{table}

The supplementary artifact verifies only the claim classes marked ZIP-runnable in its claim-to-artifact ledger. Embeddings, frozen router predictions, and router-training code are not distributed; neither end-to-end router refitting nor all figure rendering is supported.

\end{document}